\documentclass[global,twocolumn]{article}
\usepackage{graphicx}
\usepackage{url}
\usepackage{algorithm}
\usepackage{algorithmic}
\begin{document}
\title{Evolvable Agents, a Fine Grained Approach for Distributed Evolutionary Computing: Walking towards the Peer-to-Peer Computing Frontiers}
\author{J.L.J. Laredo\thanks{Department of Architecture and Computer Technology\\ETSIIT. University of Granada.}
 \and P.A. Castillo \and A.M. Mora \and J.J. Merelo
}
\date{}                     
%
%
%

%
\maketitle
\begin{abstract}

In this work we propose a fine grained approach with self-adaptive migration rate for distributed evolutionary computation. 
Our target is to gain some insights on the effects caused by communication when the algorithm scales. To this end, 
we consider a set of basic topologies in order to avoid the overlapping of algorithmic effects between communication and topological structures. We analyse the approach viability by comparing 
how solution quality and algorithm speed change when the number of processors increases and compare it with an Island model based implementation.
A finer-grained approach implies a better chance of achieving a larger scalable system; such a feature is crucial concerning large-scale parallel
architectures such as Peer-to-Peer systems. In order to check scalability, we perform a threefold experimental evaluation of this model:
First, we concentrate on the algorithmic results when the problem scales up to eight nodes in comparison with how it does following the Island model. Second, we analyse the computing time speedup of the
approach while scaling.
 Finally, we analyse the network performance with the
proposed self-adaptive migration rate policy that depends on the link
latency and bandwidth.
With this experimental setup, our approach shows better scalability than the Island model and a equivalent robustness
on the average of the three test functions under study.
\end{abstract}

\section*{Note:}
Cite this article as:\\
\noindent
\textbf{Laredo, J.L.J., Castillo, P.A., Mora, A.M., Merelo, J.J.. Evolvable agents, a fine grained approach for distributed evolutionary computing: walking towards the peer-to-peer computing frontiers. Soft Comput 12, 1145–1156 (2008). https://doi.org/10.1007/s00500-008-0297-9}

\section{Introduction}
\label{sec:intro}
Distributed applications are emerging with renewed interest since the
Peer-to-Peer (P2P) paradigm \cite{wehrle05:p2p} appeared on the area of
heterogeneous networks.  
In order to develop
applications in that environment,
questions such as \emph{scalability} (since P2P systems can become
large-scale networks) or \emph{robustness} (given that they have
dynamic topologies with continuous variants either on the connected
nodes or the network structure) become of the maximum interest and
have to be addressed \cite{spyros:robustscalable}. P2P platforms are
not only devoid of any central server, but also highly unpredictable,
since resources are added and eliminated dynamically, often as a
consequence of a decision from an user that volunteers CPUs under his
control for a certain application. Thus, within
this paradigm of volunteer computing,
EAs within P2P platforms have another issue to address:
resource allocation. 

However, regarding distributed Evolutionary Algorithms (dEAs) it is hard to find
open research lines about real scalability bounds in terms of how
many nodes could be involved within a parallel run.
What is intuitively known is
that these bounds are related to the population size and
the parallelization grain, because
 the population size represents a natural limit
i.e. the maximum number of nodes could not exceed the number
of individuals within a population; but, on the other hand, the finer the parallelization grain the
larger the number of nodes within an experiment could become.

Within the current status of research, large-scale dEAs are 
a very ambitious goal which remains out of the scope of this work.
However, we try to walk a step toward such a possibility by proposing
a fine grained agent-based approach as an alternative to the classical
coarse grained models; each individual in a population becomes an {\em
  agent} in the sense that all decisions are self-managed, not a
consequence of centrally-run processes.

The inherent parallelism of EAs \cite{eiben:eas} has been widely studied in classical existing models (see e.g. 
\cite{cantu:parallelga} for a survey) but mainly
under two approaches:  master-slave and islands. In the master-slave mode, the
algorithm runs on the master and the individuals are sent for evaluation
to the slaves, in an approach usually called {\em farming}. Using the Island model several EAs (islands) 
are used processing their own population, and exchanging the best individuals 
between islands with a certain rate and frequency \cite{cantu99:topologies}.
Both cases present major adoption problems 
in heterogeneous fully decentralized networks such as P2P networks. 
Master-slave features do not match with large-scale system robustness 
(master represents a bottleneck and a single point of failure)
and scalability (since it depends on evaluation function cost, and has
a bottleneck in the efficiency of the master performing the
evolutionary operations).
Otherwise, P2P systems
do not provide the knowledge of the global environment that the Island model
would need in order to set parameters such as the number of islands, 
the population size per island and the migration rate. 

Nevertheless, there is a third, finer grained approach, termed Fully Distributed 
model, in which processors host single individuals that evolve on their own \cite{tomassini}. 
Operations that require more than a single individual (e.g., selection
and crossover) take place among a defined set of neighbours (between individuals
on different nodes or locally available to a node). 
This model is able to adapt to heterogeneous networks since some P2P 
overlay networks 
\cite{overlaynetworks} provide a dynamic neighbourhood whose size grows
logarithmically with respect to the total size of the system in a small-world 
fashion. Following a gossip style, these small-world networks 
spread information in an epidemic manner through the whole 
network (as can be seen in \cite{jelasity:newscast,jelasity:gossip}),
which means that the risk of having obsolete individuals across the
network is minimized as a consequence of the probabilistic global 
``infection'' that the nodes undergo. 

It is obviously not straightforward to outline a method that takes
advantage of those P2P properties, obtaining at the same time high
performance and good scalability. That is why we propose a model where
each individual in an evolutionary computation population schedules
its own actions and self-adapts migration rates to the current status of the network.
The bandwidth and latency of the network links are continuously checked within every migration event. 
Hence, self-adaptation consists in increasing the migration rate when the quality of a link is high while reducing
it if it is low.
This model, which we introduce in this paper, is a step 
towards a ``Fully Distributed model'' for designing EAs in    
heterogeneous networks. 

Therefore, the main objective of this work is to provide an empirical assessment of our agent-based dEA approach
(so-called Evolvable Agent model).
To this end, we perform three case studies. The first one compares its algorithmic performance versus an island-based implementation
while scaling up to eight nodes. The second one analyses the computing time speedup within the same scenario.
Finally, we analyse the network performance regarding the self-adaptive migration rate mechanism.

The rest of the work is organized as follows: Section \ref{sec:work} presents an overview about works related to P2P Computing and dEAs. Our proposal is described in Section \ref{sec:model}, from the overall architecture of the model to a more detailed view of the specific components.
Section \ref{sec:methods} presents the methodology, experiments and parameters that we have chosen in order to assess our approach as a valid dEA. 
Section \ref{sec:result} shows and analyses the experimental results.
Finally, some conclusions are reached in Section \ref{sec:conclusions} and some future work lines proposed.

\section{State of the art}
\label{sec:work}

As it has been said, there is little research in dEAs over a P2P network;
however, each component has been tackled separately. For instance, P2P
distributed computing systems has been developed in frameworks such as {\em DREAM} \cite{arenas:dream}, which focuses in distributed processing of 
EAs and uses the P2P network DRM (Distributed Resource
Machine), {\em G2DGA} \cite{g2dga}, equivalent to the previous, but using G2P2P and
{\em JADE\footnote{Available from
\url{http://jade.cselt.it/}. Accessed on January 2007}} (Java Agent Development Framework), a P2P system which includes agents as software components.  

DRM is an implementation of the newscast protocol
\cite{jelasity:newscast}, which has served as a guide for the
proposed communication mechanism within this work.  Newscast is an
epidemic approach where every node shares local information with its
neighbourhood by selecting a node from it with uniform probability
after certain time (refresh rate).  However, our model considers a dynamic refresh
rate which depends on the quality of service (QoS) parameters latency
and bandwidth.

There are many ways of using these distributed resources in a P2P
network, but one of them is via so-called {\em agents}, which are
essentially autonomous programs that can take {\em decisions} based on
their own history and the environment. 
  Agents have been widely used in evolutionary computation: for
instance, 
Vacher et al. present in \cite{multiagent} a multi-agent approach to solve 
multi-objective problems and describe the implementation of 
functions and operators of the system.

\begin{figure*}[!htbp]
\centering
\resizebox{0.8\textwidth}{!}{%
  \includegraphics{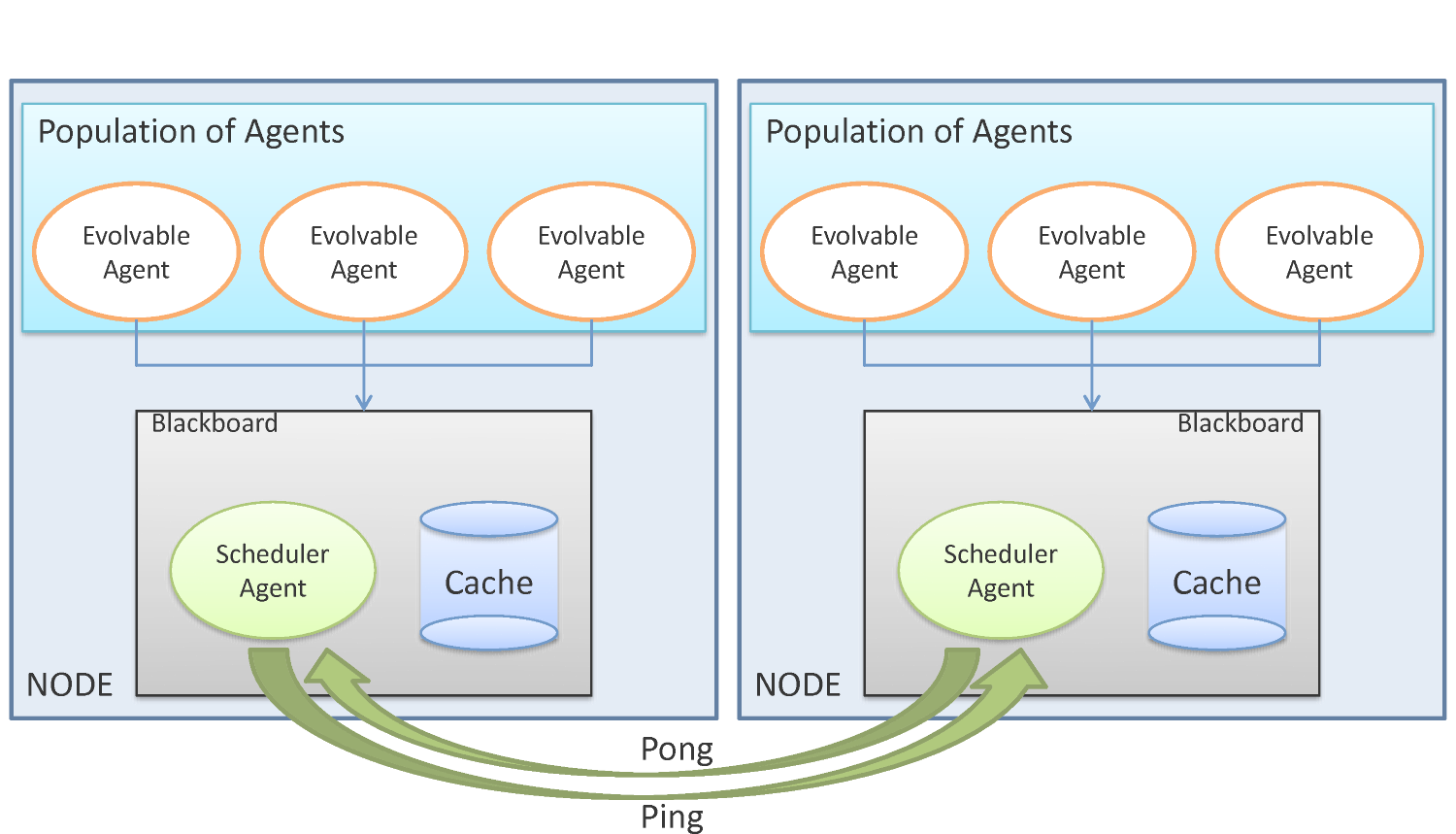}
}
\caption{Overall architecture of the model}
\label{fig:architecture}       
\end{figure*}

Some other papers attack the problem of parallel distributed
evolutionary algorithms in a wider sense:  Viveros and Bar\'an \cite{agcombinados} propose the combination of 
parallel evolutionary algorithms with local optimization functions which 
depends on processor capacities in heterogeneous computational systems. 
In some other related works on this field, Eiben et al. 
\cite{eiben:onthefly} 
report the benefits of considering population size 
adjustment on runtime. We present in  
\cite{laredo:autoadaptacion}
an agent-based system in which the load
 of every evolutionary computation experiment is self-adaptive depending 
on the architecture where it is executed, yielding more efficient results 
than the classical sequential approach.

One of the most usual and widely studied approaches in parallel EAs is the Island model (i.e. see \cite{tomassini} for a survey). The idea
behind this model is that the global panmictic population is split in several sub-populations or demes called islands.
The communication pipes between islands are defined by a given
topology, through which they exchange individuals (migrants) with a certain rate. The migration will follow a selection policy in the source island and a replacement policy in the target one.
Practitioners use to establish a fixed population size $P$ in scale-up studies, a number of islands $N$
and a population size per island of $P/n$ where $n = 1,\dots,N$. 
The work by Hidalgo and Fern\'andez \cite{Fernandez:balancing} requires a special attention. They experimentally show how the algorithmic results are strongly dependent on the number of islands. Our experimentation with the Island model is consistent with this conclusion since it also shows such a dependency. 

Finally, most of the works regarding finer grained approaches for parallel EAs focus on the algorithmic effects of using different topologies  i.e. 
Giacobini et al. study the impact of different neighbourhood structures on the selection pressure in regular lattices \cite{giacobini:regular} and different graph structures such as a toroid \cite{giacobini:gecco04} or small-world \cite{giacobini:gecco05}. This last structure has shown empirically to be competitive against panmictic EAs \cite{giacobini:evocop06}.

In this paper, we study a fine grained approach and focus on the communication issue rather than in the
topological properties. This model for distributed evolutionary
algorithms presents an 
asynchronous communication method that allows self-adaptation of the migration rate depending
on the network scenario, such feature is promising in dynamic environments such as P2P systems.
The results are matched against the classical Island model \cite{cantu:parallelga}.

\section{Overall Model Description}
\label{sec:model}

The overall architecture of our Evolvable Agent model is depicted on Figure \ref{fig:architecture}.  
It consists of a group of Evolvable Agents (each one running on its
own thread) whose main design objective is to carry out the principal steps of 
evolutionary computation: selection and variation (crossover and mutation).
In our model the selection takes place locally to each agent. Crossover and
mutation never involve many individuals, but selection in EAs usually requires 
a comparison among all individuals in the population (i.e. roulette wheel or
rank-based selection).

\subsection{Blackboard}
\label{sec:blackboard}

The main functionality of the blackboard is to keep the status of the global evolution process 
within a distributed experiment (i.e. best fitness, average fitness or the global number of  evaluations).
In addition, it also keeps a reference to any of the objects within a node which allows the interchange of
information between agents (Agent to Agent) or with cache
(Agent to Cache). The Agent to Agent communication defines the local population structure while the cache
sets the topology of the experiments. In principle, the model does not force a specific population structure, but we keep the same one than in the Island model for comparison. Thus,
 Agents will mate locally in a panmictic fashion and the cache will keep a complete graph topology (as the one sets for islands). 

\begin{figure}
\resizebox{0.5\textwidth}{!}{%
  \includegraphics{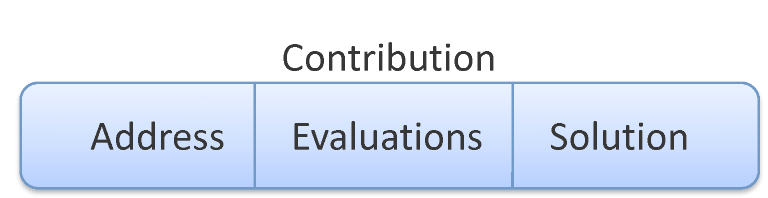}
}
\caption{Format of a cache entry. It provides the following information about 
a foreign node: Address, number of evaluations performed and one 
individual of its population termed solution}
\label{fig:contribution}       
\end{figure}

Furthermore, the blackboard implements a Scheduler (Section \ref{sec:scheduler})
that allows the spread of information among nodes in a gossip style.
The messages used among nodes are called {\em contributions} and their structure
matches with a cache entry (Figure \ref{fig:contribution}).  Thus,
instead of the classical view of a population of solutions managed by an oracle, 
this model proposes a population of autonomous agents, each one representing a solution.

\subsection{Scheduler Agent and Cache}
\label{sec:scheduler}

Algorithms \ref{alg1}, \ref{alg2} and \ref{alg3} show the pseudo-code of the 
main tasks in the 
communication process that builds the overlay network. Each blackboard 
maintains a cache with a maximum of one entry per
node in the network. Each entry follows the contribution format (Figure 
\ref{fig:contribution}) in which the Address field is an univocal index for nodes.
This way, the cache replaces the oldest entries from a given node using the newest contributions. This process 
leads to the removal of obsolete individuals and allows a global evolution in a decentralized environment.

\begin{algorithm}
\caption{Scheduler Agent}
\label{alg1}
\begin{algorithmic}

\STATE $\Delta T \Leftarrow$ 1 sec.
\LOOP
\STATE sleep $\Delta T$
\STATE Node $\Leftarrow$ Random selected node
\STATE Sol $\Leftarrow$ Selected Solution in $P_{agents}$
\STATE Contribution $\Leftarrow$ $Num_{Evaluations}$,Sol
\STATE Ping (Node,Contribution)
\ENDLOOP
\end{algorithmic}
\end{algorithm}

\begin{algorithm}
\caption{Ping Handler}
\label{alg2}
\begin{algorithmic}

\REQUIRE Node, Contribution
\STATE Cache(Node)$\Leftarrow$ Contribution
\STATE Pong(Node,OK)

\end{algorithmic}
\end{algorithm}
\begin{algorithm}
\caption{Pong Handler}
\label{alg3}
\begin{algorithmic}

\REQUIRE Node
\STATE $\Delta T \Leftarrow$ Time used answering the Ping

\end{algorithmic}
\end{algorithm}

 The scheduling mechanism is carried out by each node as explained next:

\begin{itemize}
\item \textbf{Algorithm \ref{alg1}:} After $\Delta T$ time, the current 
node selects another node from cache with uniform probability to establish 
communication. Current node sends an application level $Ping$ message to 
the selected node with information about a random solution in the population of
 agents ($P_{agents}$) in a contribution format (Figure \ref{fig:contribution}).
\item \textbf{Algorithm \ref{alg2}:} The selected node stores that solution
in its cache and sends back an acknowledgement message ($Pong$).
\item \textbf{Algorithm \ref{alg3}:} At the arrival of the $Pong$, the current node 
updates its refresh rate ($\Delta T$) with the time spent in the operation.
\end{itemize}

Next we present a possible method for an Evolvable Agent where selection is locally executable,
in the sense that it does not require direct comparison with all individuals in the population.

\subsection{Evolvable Agent with Tournament Selection}
\label{sec:tournament}

Algorithm \ref{alg4} shows the pseudo-code of an Evolvable Agent that uses
Tournament Selection.
The agent owns a solution ($S_{t}$) which it tries to evolve. 
The selection mechanism works as follows:
Each agent selects $k$ ($k$ = tournament size) solutions 
from other agents and solutions stored in cache (which are migrants from 
network nodes) with uniform probability using the blackboard.
The two best solutions are stored in $Sols$ ready to be recombined by
a crossover operator. The crossover returns a single solution $S_{t+1}$ that
is mutated and evaluated. If the newly generated solution $S_{t+1}$ is better than
the old one $S_t$, it becomes the current solution. 
Finally, Blackboard maintains global elitism by storing the best solution found so far
in $Blackboard.BestSol$.

\begin{algorithm}
\caption{Evolvable Agent with Tournament Selection}
\label{alg4}
\begin{algorithmic}

\STATE $S_{t}$ $\Leftarrow$ Initialize Agent
\STATE Register Agent on the blackboard

\LOOP
\STATE Sols  $\Leftarrow$ Selection($k$, Blackboard)
\STATE $S_{t+1}$ $\Leftarrow$ Recombine(Sols,$P_{c}$) 
\STATE $S_{t+1}$ $\Leftarrow$ Mutate($S_{t+1}$, $P_{m}$)
\STATE $S_{t+1}$ $\Leftarrow$ Evaluate($S_{t+1}$) 
\IF{$S_{t+1}$ better than Blackboard.BestSol}
\STATE    Blackboard.BestSol $\Leftarrow$ $S_{t+1}$ 
\ENDIF
\IF{ $S_{t+1}$ better than $S_{t}$} 
\STATE    $S_{t}$ $\Leftarrow$ $S_{t+1}$
\ENDIF
\ENDLOOP
\end{algorithmic}
\end{algorithm}

\section{Methods and Experiments}
\label{sec:methods}

We have carried out an empirical investigation over the Evolvable Agent model.
It consists of three case studies. Firstly, we compare our proposed approach with
the Island model in a real parallel scenario with up to eight nodes. Secondly, we show the improvement on computing time 
while the algorithm scale up to those eight nodes. Finally,  we evaluate the network performance
under the proposed self-adaptive migration rate (see section \ref{sec:scheduler}) and different migration rates for the
Island model. 

\subsection{The Benchmark}
\label{sec:benchmark}

We have chosen as a benchmark three real-coding test functions from the test suite by Suganthan et al. \cite{CEC2005RealParameterOptimization} used in the 2005 IEEE Congress on Evolutionary Computing (CEC'05). This set includes
one unimodal function derived from a sphere and two multimodal functions: Rastrigin and Schwefel.

It is important to note that our research objective is not to 
outperform existing results. Rather, these functions are used only as
test problems for the empirical study among different proposed
algorithms.

\subsubsection{Shifted Sphere function.}
\label{sec:sphere}

\begin{figure}[!htpb]
\resizebox{0.5\textwidth}{!}{%
  \includegraphics{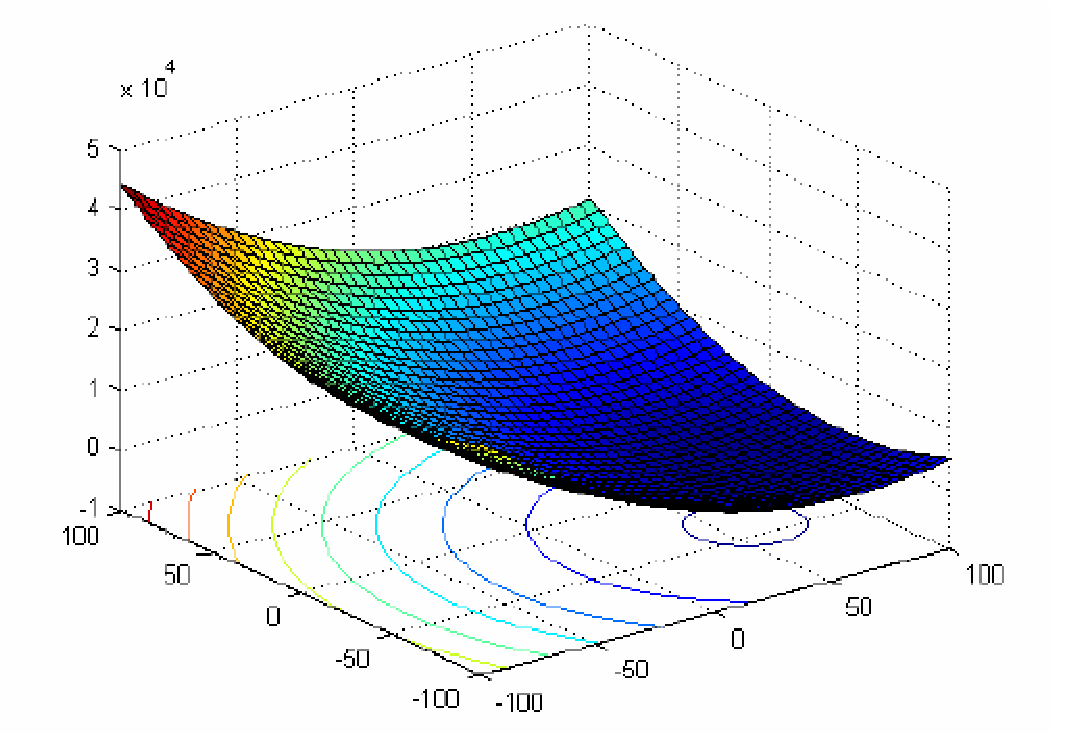}
}
\caption{Sphere function in two dimensions (Suganthan et al., 2005)}
\label{fig:sphere}       
\end{figure}

The {\em shifted sphere} function  is an
unimodal, scalable function;  Figure \ref{fig:sphere} represents the shape
of the two-dimensional version of the function; its formula is

\begin{equation} \label{eq:sphere}
 F(x)=\sum_{i=1}^D z_i^2 +f_{bias} 
\end{equation}
\[\overline{z}=\overline{x}-\overline{o} \]

where $D$ represents the number of dimensions, $-100 \leq x_i \leq 100$ and $o=[o_1,o_2,\dots,o_D]$ the shifted 
global optimum. The optimum (minimum) is $f_{bias}=-450$. We have set $D=100$

\subsubsection{Shifted Rotated Rastrigin's function.}
\label{sec:rastrigin}

\begin{figure}[!htpb]
\resizebox{0.5\textwidth}{!}{%
  \includegraphics{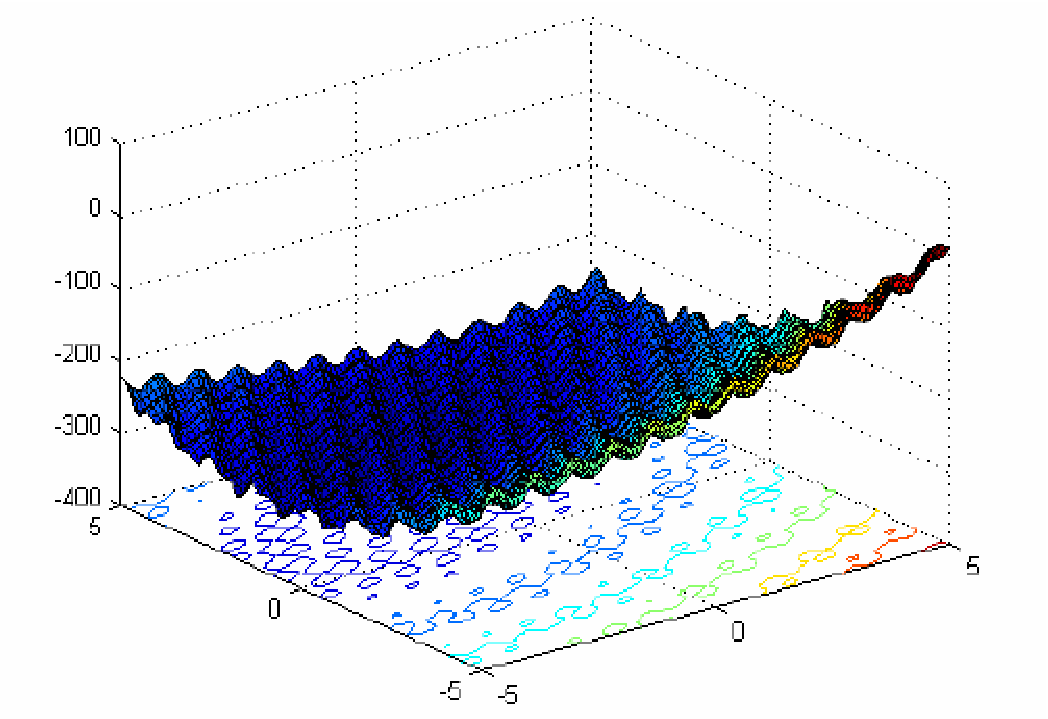}
}
\caption{Rastrigin function (Suganthan et al., 2005)}
\label{fig:rastrigin}       
\end{figure}

The shifted rotated Rastrigin's function is a multimodal function with
a huge number of local optima, Figure \ref{fig:rastrigin} represents
the shape of the two-dimensional function, and its definition is

\begin{equation} \label{eq:rastrigin}
F(x)= \sum_{i=1}^D ( z_i^2 - 10 cos (2 \pi z_i) + 10 ) + f_{bias} 
\end{equation}

where $ -5 \leq x_i \leq 5 $ and the global optimum is
$f_{bias}=-330$. In this paper we have used $D=30$.

\subsubsection{Schwefel function.}
\label{sec:schwefel}

\begin{figure}[!htpb]
\resizebox{0.5\textwidth}{!}{%
  \includegraphics{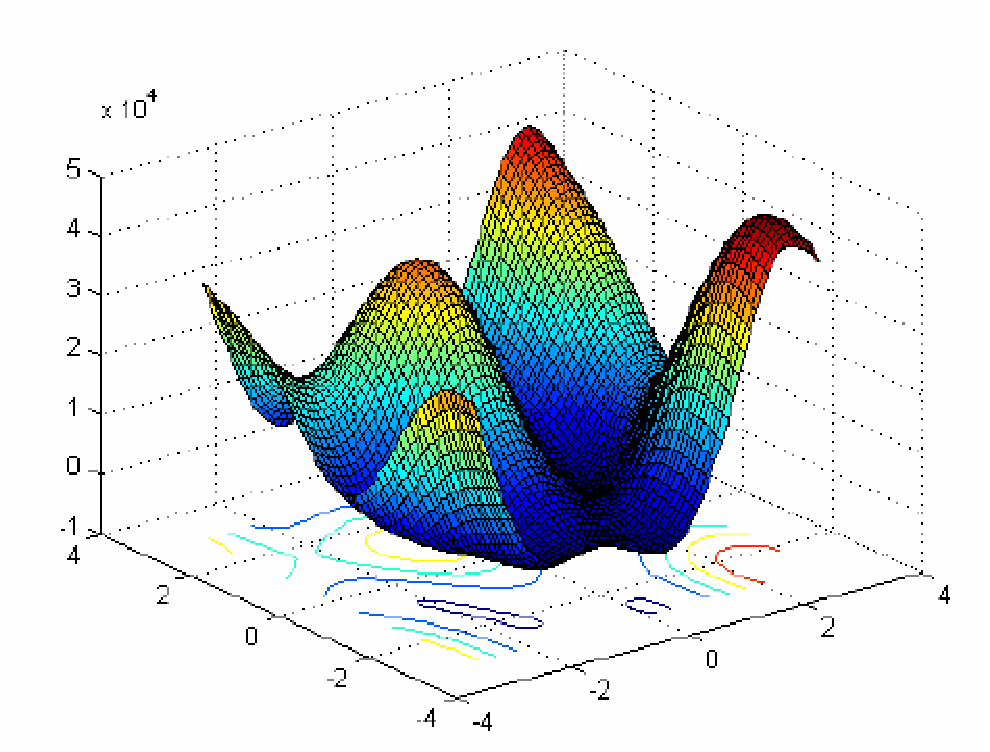}
}
\caption{Schwefel function (Suganthan et al., 2005)}
\label{fig:schwefel}       
\end{figure}

\begin{figure*}[!htpb]
\centering
\resizebox{0.8\textwidth}{!}{%
  \includegraphics{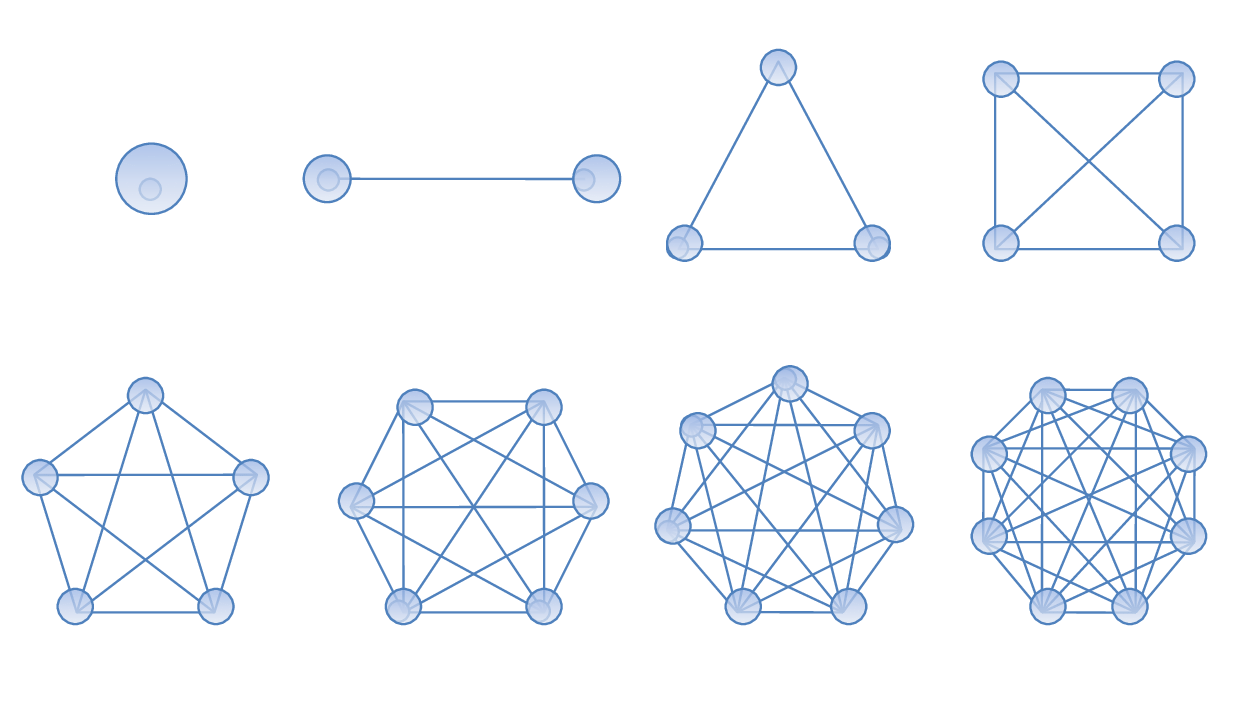}
}
\caption{Test topologies}
\label{fig:complete}       
\end{figure*}

The Schwefel's function is another multimodal function, whose
two-dimensional version is represented in  Figure
\ref{fig:schwefel}, and whose definition is

\begin{equation} \label{eq:schwefel}
F(x)= \sum_{i=1}^D (A_i - B_i(x))^2 + f_{bias} 
\end{equation}
\[A_i = \sum_{j=1}^D (a_{ij}sin(\alpha_j) + b_{ij}cos(\alpha_j)) \]
\[B_i(x) = \sum_{j=1}^D (a_{ij}sin(x_j) + b_{ij}cos(x_j)) \]

where $ -\pi \leq x_i,\alpha_i \leq \pi $, $a_{ij},b_{ij} \in
\{-100,100\}$ and the global optimum is $f_{bias}=-460$. In this paper
we have used $D=10$.

\subsection{Experimental Setup}
\label{sec:setup}

All EA variants share the main evolutionary operators and rates. All results are the estimation of 30 independent runs per experiment. Next settings  remain invariant during the whole experimentation:

\begin{itemize}
\item Initialization of the population: Random.
\item Stopping criteria: The computational effort is fixed at 2500000 evaluations.
\item Operators: Uniform crossover and mutation, Tournament selection
  with tournament size = 2.
\item Elitism.
\item Crossover probability: 0.9.
\item Mutation probability: 0.01.
\item Population size: 512.
\end{itemize}

The settings for running the Island model are:

\begin{itemize}
\item Evolutionary scheme: Generational with elitism
\item Migration frequency: We have established several migration frequencies. The rate is one individual per migration each 25, 50, 75 or 100 generations. 
\item Migration policies: Best individuals are selected for migration and randomly replaced in the target island.
\end{itemize}

The objective of this work is to provide experimental data on how solution quality and
algorithm speed change with the number of processors. Therefore, we compare results obtained 
on a single node up to eight nodes. The physical test-bed for the parallel scenario is shown in
Table \ref{table:parallel}.

\begin{table}[h]
\centering
\begin{tabular}{|c|c|}
\hline
\texttt{Test-bed}&\\
\hline
Number of Nodes&8\\
Node Processor& 2X AMD Athlon(tm)2400+\\
Network& Gigabit Ethernet\\
Operating System & Linux (kernel 2.6.16-1.2096\_FC5smp)\\
Java version & J2RE (version 1.5.0\_06)\\
\hline
\end{tabular}
\caption{Parallel scenario test-bed
\label{table:parallel}}
\end{table}

For all experiments with $n$ nodes ($n = 1, \dots, 8$) we used a fully connected graph topology as depicted in Figure \ref{fig:complete}.
The complexity grows with an order of $O(\frac{(n-1)n}{2})$.  
Such scenario represents the worst case in network topologies since it intensifies the impact of communication overhead. A real P2P 
overlay network should grow with a smaller order of complexity i.e. following a
small-world fashion. As designed, the Scheduler Agent (section \ref{sec:scheduler}) will self-adapt the
refresh rate $\Delta T$ to the congestion of links. 

Load balancing is performed in a straightforward way:
\begin{itemize}
\item Population is distributed equally among nodes in each
  experiment. Therefore, the size of subpopulations is 512 individuals/$n$  
where $n=1,\dots,8$ nodes.
\item A node selects randomly a neighbour for migration.
\item The communication process takes place asynchronously in both models.
\end{itemize}

Finally, software is available under GNU public license on the repository \url{https://forja.rediris.es/projects/geneura/}.

\section{Experimental Results}
\label{sec:result}

Any approach towards distributed EAs in heterogeneous 
networks has to deal at least with two main issues. First, it has to avoid altering the algorithmic 
results as a consequence of scaling up to certain number of nodes. Unfortunately, as shown in \cite{Fernandez:balancing},
balancing the computational effort in several subpopulations has an implicit algorithmic effect. Second, the use of
more computational power has to report time profits.

In order to tackle such issues, we have studied our model under three perspectives: 
\begin{itemize}
\item \textbf{Case study 1: Evolvable Agents vs. Island model.} This case study
concentrates on the algorithmic results when the problem scales up to eight nodes in comparison with how it does following
the Island model. 
\item \textbf{Case study 2: Time performance in a Parallel Scenario.} We analyse the runtime performance of our approach in a
parallel scenario up to eight nodes.
\item \textbf{Case study 3: Network performance.} Finally, we analyse the network performance. Our proposal follows the idea that
the better the QoS of links the closer the evolution could be to a panmictic reproduction. To this end we proposed
the self-adaptive migration rate exposed in section \ref{sec:scheduler}. 
\end{itemize}

The results obtained in these case studies are presented in turn in
the next subsections.

\subsection{Case Study 1: Evolvable Agents vs. Island model}
\label{sec:study1}

This case study focuses on the analysis of the scaling properties of
the  Evolvable Agent vs. the Island model up to eight nodes. From
the results we check that the different migration rates do not seem to affect the scalability performance in the Island model.
A pairwise t-Student analysis among them reveals that there are no
significant differences on the distributions. Hence, for the sake of
clarity, we just show within this case study the experiments with a
migration rate of once every 25 generations.

\subsubsection{Shifted Sphere function.}
\begin{figure*}[!htbp]
\centering
\resizebox{1\textwidth}{!}{%
  \includegraphics{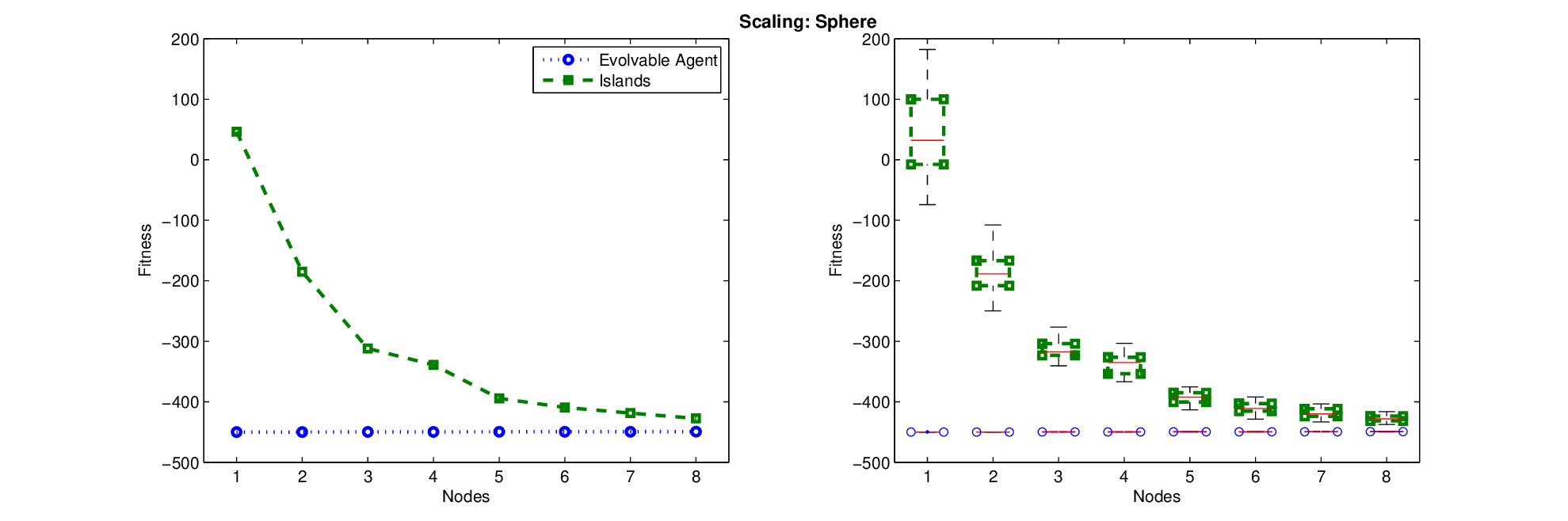}
}
\caption{Mean ({\em left}) and box-plot representation ({\em right}) of the best-fitness  for the sphere problem. Evolvable Agent vs. Island model in a parallel scenario from 1 to 8 nodes.}
\label{fig:algorsphereadaptive}       
\end{figure*}
\begin{figure*}[!htbp]
\centering
\resizebox{1\textwidth}{!}{%
  \includegraphics{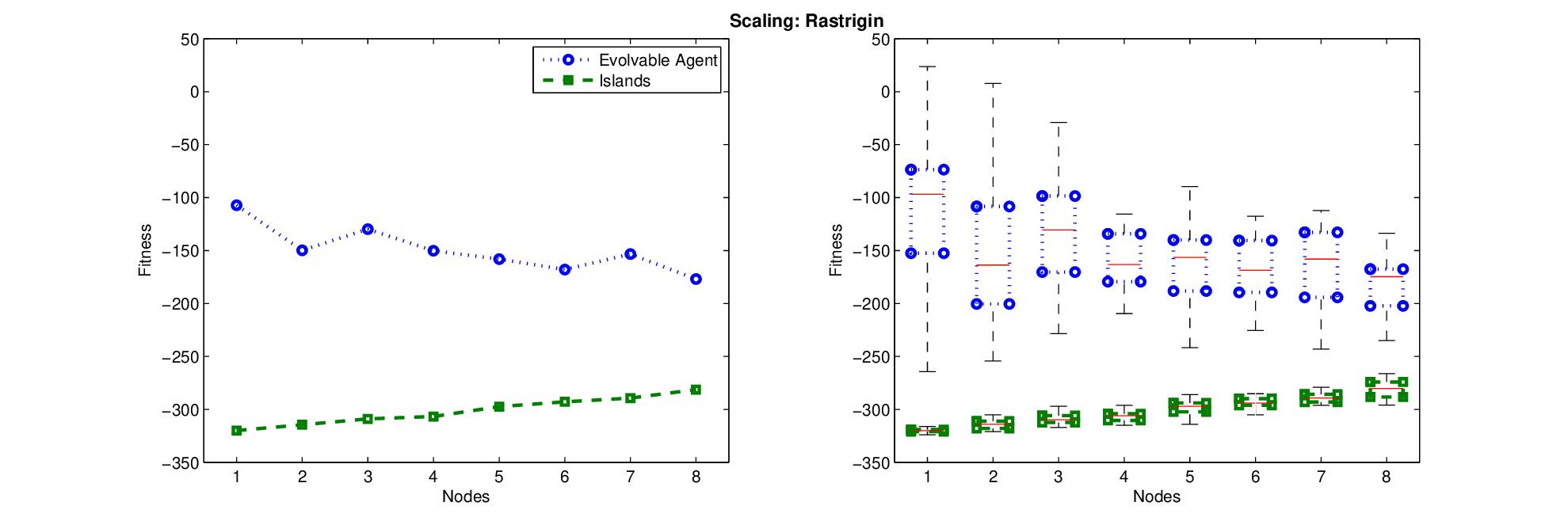}
}
\caption{Mean ({\em left}) and box-plot representation ({\em right}) of the best-fitness for the Rastrigin's problem. Evolvable Agent vs. Island model in a parallel scenario from 1 to 8 nodes.}\label{fig:algorrastriginadaptive}       
\end{figure*}
\begin{figure*}[!htbp]
\centering
\resizebox{1\textwidth}{!}{%
  \includegraphics{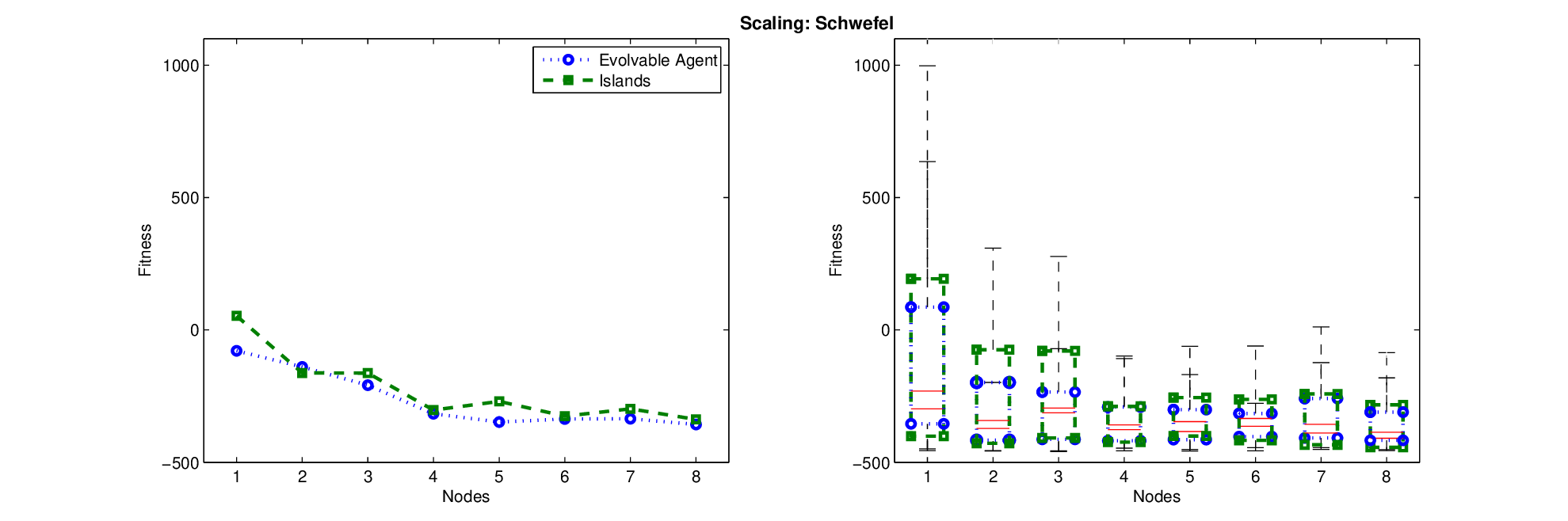}
}
\caption{Mean ({\em left}) and box-plot representation ({\em right}) of the best-fitness for the Schwefel's problem. Evolvable Agent vs. Island model in a parallel scenario from 1 to 8 nodes.}\label{fig:algorschwefeladaptive}       
\end{figure*}

The Figure \ref{fig:algorsphereadaptive} depicts the algorithmic performance of the
Evolvable Agent model against the Island model while scaling from 1 to 8 nodes.

From the analysis of the graph we can see that:

\begin{itemize}
\item The Evolvable Agent model yields better results than the Island model for this unimodal function.
\item The problem optimum is -450 and all results in the Evolvable Agent case are behind -448.2 as shown in Figure \ref{fig:algorsphereadaptive}. Meanwhile, the Island model yields results from almost 200 in the worst case (1 node) and improves as the number of nodes increase (number of nodes and demes match within our experimentation). 
\item The scale up analysis shows a more robust performance of the Evolvable Agent with a low variation in the results against the big differences in the case of the Island model. 
\item As it can be seen, the standard deviation of the solutions is much
smaller for the Evolvable Agent than for the Island model which means
that, when increasing the number of nodes, the dispersion of the
fitness obtained after a common number of evaluations is smaller.
\end{itemize}

\subsubsection{Shifted Rotated Rastrigin's function.}

Figure \ref{fig:algorrastriginadaptive} depicts the algorithmic scale
up performance of the Evolvable Agent model and the Island model. The
behavior observed here is radically different from the one shown above.

\begin{itemize}
\item The Island model yields better and more robust results than the Evolvable Agent model.
\item The best-fitness distributions of the Island model get worse while scaling. 
\item However, in spite of a higher standard deviation than the Island
  model, the Evolvable Agent model does not present significant
  differences for the distributions from 4 to 7 nodes. Therefore,
  although less robust for this function than the generational scheme,
  our model is robust against scaling since it yields stable results.
\end{itemize}

The results in this case also show that the parameters that yield the
best result in the Island model might not be the same as for the
Evolvable Agent model, specially in the case of highly deceptive
functions such as this one. Specifically, the fact that Evolvable
Agents send a random member of its population to the rest of the
nodes, while the Island model sends the best, might contribute to its
success in this case. However, this is an inherent feature in the
Evolvable Agent model, not a choice, thus, taking this into account,
some other measures will have to be inserted into the model in order
to make it better suited for this kind of functions.

\subsubsection{Schwefel's function.}

The algorithmic performance is equivalent for both models in the case of the Schwefel's function as can be seen in the Figure
\ref{fig:algorschwefeladaptive} where the Evolvable Agent and the Island model 
yield results close to -460 (the function optimum). Best-fitness distributions scale in a robust way keeping same results from one to eight nodes.

\subsubsection{Conclusions from Case Study 1.}
Since distributed EAs suffer structural changes at population level which
modify their algorithmic behaviour, we can conclude from the previous 
observations that our model proposal shows a robust behaviour in two of the three test functions compared with the Island model.

The Evolvable Agent model yields better results than the Island model for the sphere function, worse for the Rastrigin's one
and equivalent for the Schwefel function.  These results over three test functions are consistent with the "no free lunch" 
theorem \cite{wolpert97no} where the performance of two searching algorithms are on average the same when they are tested over all possible functions.



\subsection{Case Study 2: Time performance in a Parallel Scenario}
\label{sec:study2}

\begin{figure}[!htbp]
\resizebox{0.5\textwidth}{!}{%
  \includegraphics{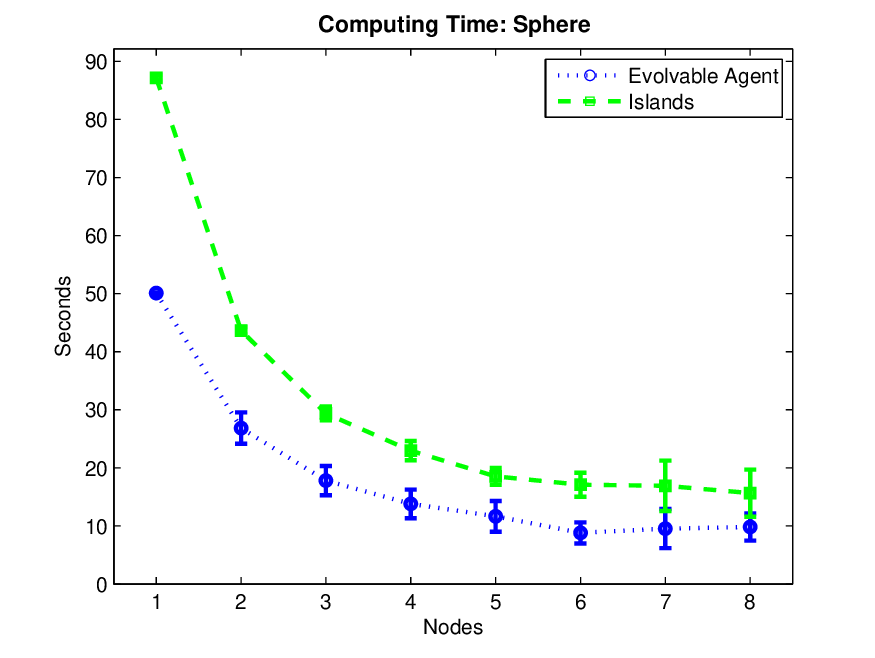}
}
\caption{Average computing time and its respective standard deviation up to 8 nodes for the Sphere function. Islands vs. Evolvable Agent.}
\label{fig:timesphere}       
\end{figure}

\begin{figure}[!htbp]
\resizebox{0.5\textwidth}{!}{%
  \includegraphics{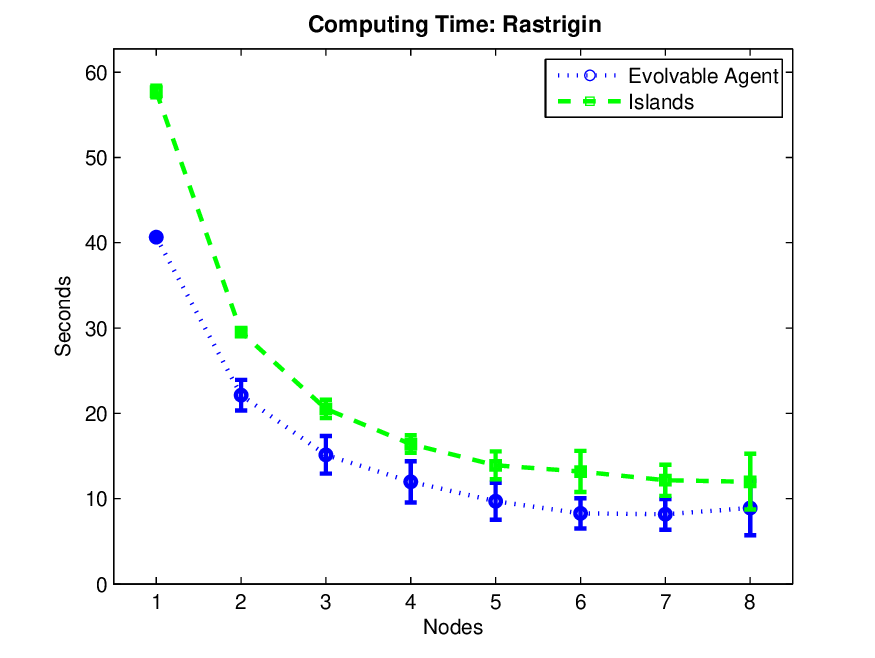}
}
\caption{Average computing time and its respective standard deviation up to 8 nodes for the Rastrigin's function. Islands vs. Evolvable Agent.}
\label{fig:timerastrigin}       
\end{figure}

\begin{figure}[!htbp]
\resizebox{0.5\textwidth}{!}{%
  \includegraphics{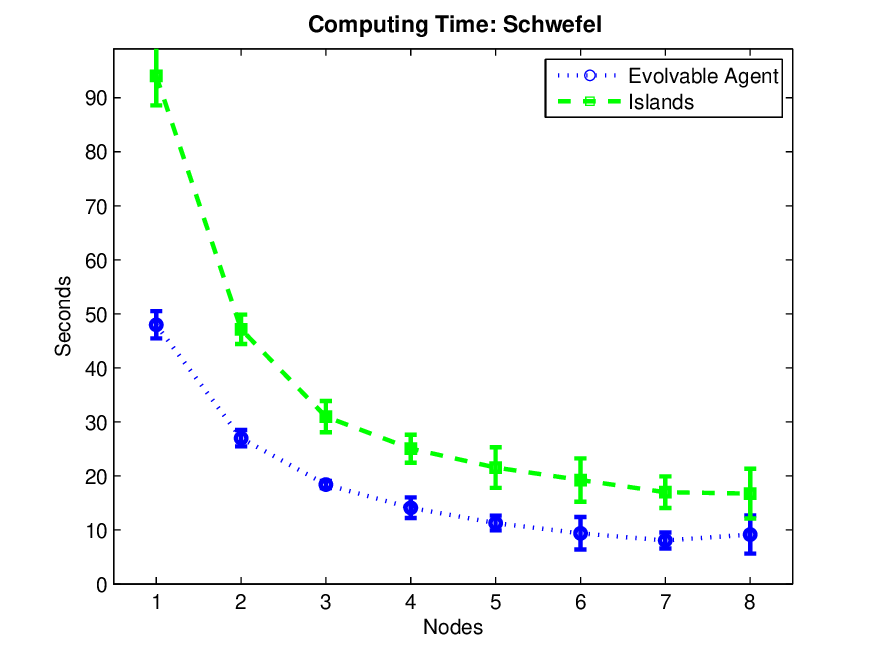}
}
\caption{Average computing time and its respective standard deviation  up to 8 nodes for the Schwefel's function. Islands vs. Evolvable Agent.}
\label{fig:timeschwefel}       
\end{figure}

Figures \ref{fig:timesphere}, \ref{fig:timerastrigin} and \ref{fig:timeschwefel} show the time performance for the
different test functions when they scale from one to eight nodes. 
Evolvable Agent and Island models yield a good speed-up which scales in a linear fashion.

Nevertheless, as can be seen, the time performance of the Evolvable
Agent is better than the Island model independently either of the
number of nodes or the test functions. That is due to the single
process nature of the islands against the thread-based nature of the
agents. Since our test-bed is composed of biprocessor nodes (see Table \ref{table:parallel}), our approach balances the load between the two processors while a single island process can just take advantage of one of them.

\subsection{Case Study 3: Network performance }
\label{sec:study3}

Within this case study we have measured the latency of the messages (migrants). They are wrapped in a high level ping according to the asynchronous self-adaptive refresh rate exposed in section \ref{sec:scheduler}. Since the size of individuals is constant for each experiment, the bandwidth needed for each migration will be constant too. Therefore, in order to measure the network performance we focus on the average latency that a migrant uses. A higher latency indicates a heavier traffic within the network.

\begin{figure}[!htbp]

\resizebox{0.5\textwidth}{!}{%
  \includegraphics{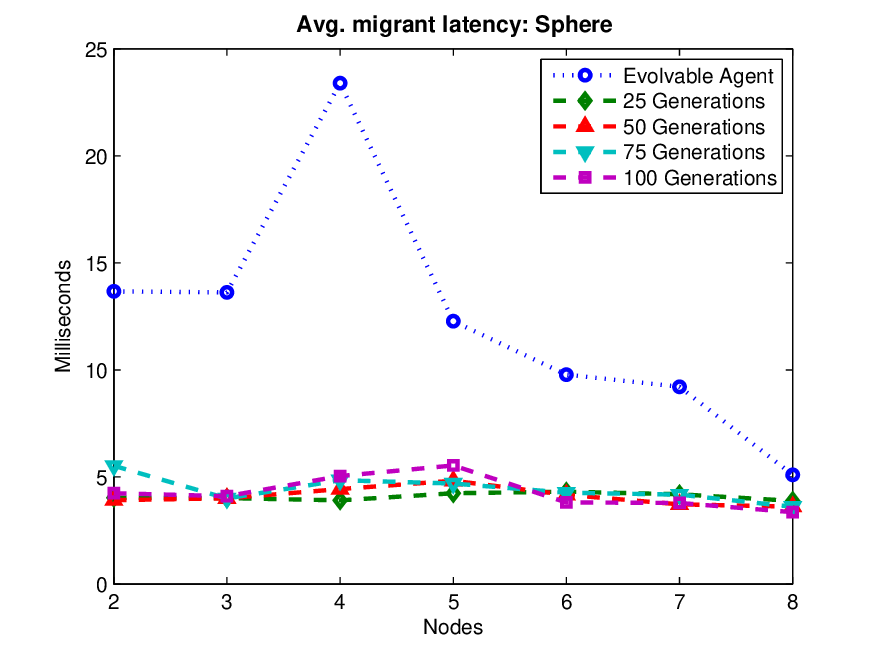}
}
\caption{Average migrant latency for the sphere problem. Each migrant is a real-coded vector of size 100. The graph shows the latency of our proposed
self-adaptive rate and the 25, 50, 75 and 100 generations frequency of the island-based implementation}
\label{fig:sph}       
\end{figure}

\begin{figure}[!htbp]

\resizebox{0.5\textwidth}{!}{%
  \includegraphics{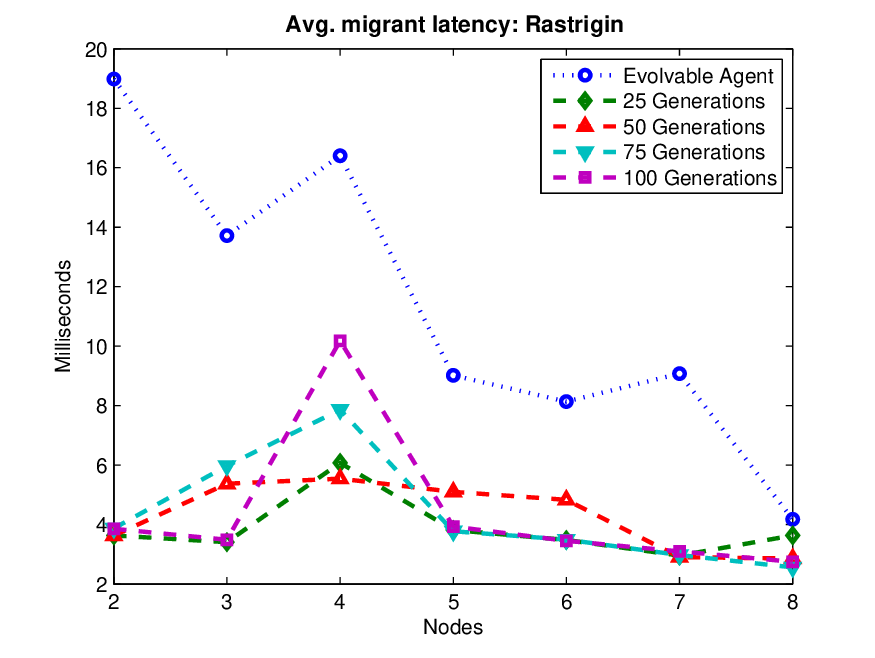}
}
\caption{Average migrant latency for the Rastrigin's problem. Each migrant is a real-coded vector of size 30. The graph shows the latency of our proposed
self-adaptive rate and the 25, 50, 75 and 100 generations frequency of the island-based implementation}
\label{fig:ras}       
\end{figure}

\begin{figure}[!htbp]

\resizebox{0.5\textwidth}{!}{%
  \includegraphics{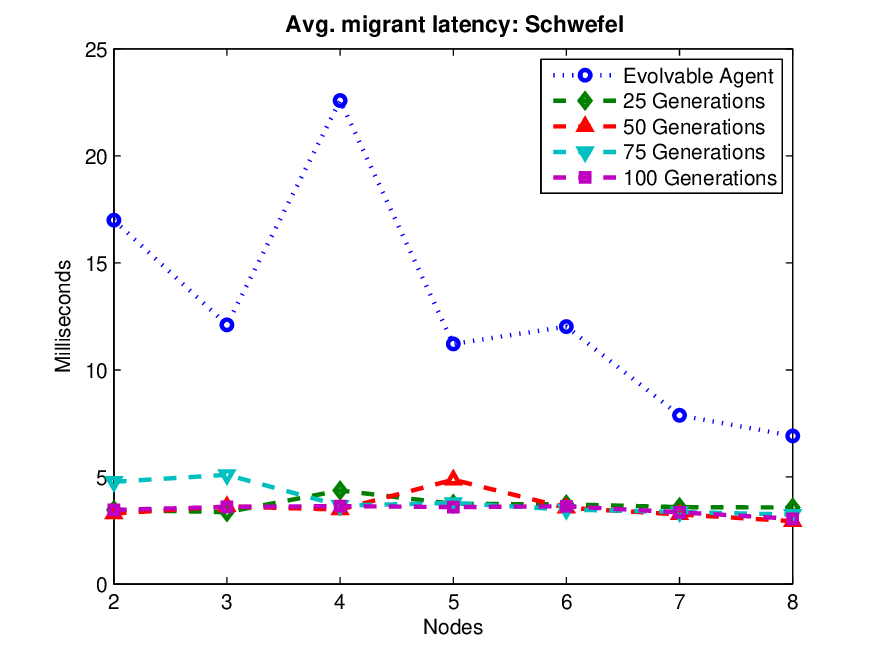}
}
\caption{Average migrant latency for the Schwefel's problem. Each migrant is a real-coded vector of size 10. The graph shows the latency of our proposed
self-adaptive rate and the 25, 50, 75 and 100 generations frequency of the island-based implementation}
\label{fig:sch}       
\end{figure}

Figures \ref{fig:sph}, \ref{fig:ras} and \ref{fig:sch} respectively depicts the average latency per migrant in the case of the sphere, Rastrigin and
Schwefel problems. Each figure shows the average latency using either the Evolvable Agent model and the Island model, this last using migration rates of 
one individual per 25, 50, 75 and 100 generations.

The graphs show that the two dEAs under study behave quite differently
with scale. The migration process in the different island settings remains almost
constant, no matter the migration rates nor the number of nodes, the values stay around 4-6 milliseconds in the sphere problem (Figure \ref{fig:sph}), 3-10 in the Rastrigin's problem (Figure \ref{fig:ras}) and 3-5 in Schwefel's one (Figure \ref{fig:sch}). Meanwhile, the average latencies in the Evolvable Agent approach do not seem to follow a clear pattern, going from 4 to 33 milliseconds.

Within all the experiments the Island model show that the migration frequency does not have a big impact on the average latency. 
Hence, the network capability would allow a higher migration rate;
however, it should be taken into account that migration takes place in
bursts, not spread uniformly along the experiment.
We can conclude that the overall use of the network is more intensive in the case of
our proposal, but, on the other hand, it is more uniformly spread. It is hard to recognize a clear pattern about how
the latency changes when the number of nodes increases. Although it seems to decrease as the complexity of the network increases showing a sort of self-adaptive response to the scenario. Future works will have to dive into a deeper analysis of the self-adaptive
migration rate. Besides, latency at this stage is not an issue since
it does not have an impact in performance; some experiments with
bigger populations and a higher number of nodes would have to be done
to check at what point it becomes an issue, if ever.

Finally, on the one hand, the case study 1 (section \ref{sec:study1})
is an experimental proof that a higher migration frequency is able to
reach a  robust behaviour in our distributed algorithm when the number
of nodes increases. 
On the other hand, case study 2 (section \ref{sec:study2}) shows how
the Evolvable Agent is able to outperform the Island model in terms of
computing time despite the increase in communications.  


\section{Conclusions and Future Works}
\label{sec:conclusions}

In this paper we have presented a fine grained approach that would be
key in an efficient fully 
distributed EA model. The model is designed to deal with the features
of large-scale networks, specially P2P networks. The 
evolution process consists in maintaining a self-organizing population
of evolving agents that represent
single solutions; each agent can access other agents' current solution in operations that needs 
more than one individual (e.g. selection) by means of the blackboard mechanism.

From the proposed experiments the following conclusions can be reached:
\begin{itemize}
\item The structural changes on the EA that 
our model proposes with respect to generational models (i.e. Island model) yield robust results as has been shown in Section \ref{sec:study1}.

\item Experimental data show that our approach speed scales well despite the growing topological complexity while maintaining solution
quality. Unfortunately, eight nodes are still a small network for extrapolating these results to large-scale networks.

\item Island model also shows a good (linear) scaling
  behavior. However, the single process nature of the islands is not
  able to take advantage of Symmetric Multiprocessing 
  architectures as the thread-based nature 
of our agent approach does. Since our test-bed is composed of biprocessor nodes, our approach outperforms the computing
time of the Island model independently of the number of nodes or the
test functions. 

\item For a pre-established computational effort, the best 
fitness distributions of our approach in the different test topologies do not show
significant differences in most cases, which means that equal
computational effort leads to the same results, with a gain in time,
since the effort is distributed. 

\item The proposed self-adaptive migration rate leverage  network
  capacity better than the fixed migration rate of the Island model 
as shown in 
section \ref{sec:study3}. Nevertheless, at this point, it is hard to recognize a clear pattern about how the latency changes when the number of nodes 
increases. Future work will focus on the study of the self-adaptive migration rate.

\item This approach is a proof of concept towards a distributed EA model where
scalability and robust results are possible together. Such features
are of the utmost importance regarding dEAs on P2P architectures.

\end{itemize}

Future work will have to consider the experimentation in large-scale
networks where further conclusions can be reached respecting
scalability limitations, fault tolerance, adaptation to heterogeneity and algorithmic
effects of having high latency links. Within this line we plan to
implement the model into a P2P framework such as DREAM
\cite{arenas:dream} which shares its main design objectives.

Finally, beyond these preliminary tests, we consider as a future line
of work checking runtime population size adjustment as a natural
mechanism for adaptation to dynamic P2P networks where available
resources are changing through the time. Within this line, our work
\cite{laredo:selection} focuses on self-adaptive population size by
the use of the Autonomous Selection Mechanism exposed in
\cite{eiben:autonomous} and \cite{upali:adaptive}.

\section*{Acknowledgments}

This work has been supported by the Spanish MICYT project
TIN2007-68083-C02-01, the Junta de Andalucia CICE project P06-TIC-02025
and the Granada University PIUGR 9/11/06 project.

%
%

\end{document}